# Encoding models for scholarly literature

*Does the TEI have a word to say?*

Martin Holmes, Humanities Computing and Media Centre, University of Victoria

Laurent Romary, INRIA-Gemo & Humboldt Universität Berlin

Abstract: In this chapter, we examine the issue of digital formats for document encoding, archiving and publishing, through the specific example of "born-digital" scholarly journal articles. This small area of electronic publishing represents a microcosm of the state of the art, and provides a good basis for this discussion. We will begin by looking at the traditional workflow of journal editing and publication, and how these practices have made the transition into the online domain. We will examine the range of different file formats in which electronic articles are currently stored and published. We will argue strongly that, despite the prevalence of binary and proprietary formats such as PDF and MS Word, XML is a far superior encoding choice for journal articles. Next, we look at the range of XML document structures (DTDs, Schemas) which are in common use for encoding journal articles, and consider some of their strengths and weaknesses. We will suggest that, despite the existence of specialized schemas intended specifically for journal articles (such as NLM), and more broadly-used publication-oriented schemas such as DocBook, there are strong arguments in favour of developing a subset or customization of the Text Encoding Initiative (TEI) schema for the purpose of journal-article encoding; TEI is already in use in a number of journal publication projects, and the scale and precision of the TEI tagset makes it particularly appropriate for encoding scholarly articles. We will outline the document structure of a TEI-encoded journal article, and look in detail at suggested markup patterns for specific features of journal articles. Next, we will look briefly at how XML-based publication systems work, and what advantages they bring over electronic publication methods based on other digital formats.

## Introduction

This book chapter provides an overview on issues related to the definition of a standard framework for the editing of scientific content. It mainly takes its examples from the specific case of journal papers, while attempting to cover the core features of similar documents (conference papers, scientific books, ISO standards, etc.). The focus on scholarly papers results from a series of converging factors indicating that the provision of a reference model for the representation of such textual objects has become a central aspect of the capacity of scholarly publishing to go digital.

These various factors may be summarised as follows:

- Most of the digital edition workflow is now carried out almost entirely in electronic form. Authors and reviewers are only exchanging digital texts with publishers;

- In the scientific world itself, the increasing role of publication repositories, in conjunction with the open access movement, has raised questions, as well as expectations, with regards long-term accessibility of the corresponding data;

- Specific repositories such as Pubmed Central[1] have even taken strong positions with regard to the kind of formats they will offer for long-term accessibility;

- XML technology has gained enough maturity to be now considered as the natural syntactic framework for the representation of semi-structured data in general, and particularly text based documents;

- Even when taking the XML technology for granted, one can observe that so far no specific XML application has emerged as a *de facto* nor *de jure* standard, and even worse, no coordinated vision seems to guide the development of ongoing initiatives.

This chapter will approach the issue from the point of view of the actual use cases and needs of an editing workflow, identifying how the various types of workflows (author - publisher (reviewer) - reader), the issues and constraints related to scholarly publishing (what is specific to journal papers as opposed to any kind of semi-structured document), and style guides for scientific publications may impact on the definition of a reference model and/or format. In this context, we will try to demonstrate how much one has to consider the representation of scholarly papers in the wider context of text representation, in order to provide both a wide and sound basis for standardization but also to ensure a long-term convergence between specific and generic document types, through the reuse of shared components. This will lead us to suggest that the Text Encoding Initiative can be a good candidate to depart from proprietary endeavours and we will try to characterize a TEI subset for journal editing that covers most of the features identified in our paper.

## Scholarly publishing and open-access

It would be quite difficult to address the domain of scholarly publishing from the academic viewpoint without tackling, at least partially, the open access debate. To make a long story short, the open access debate is rooted in the serial crisis that took place in the 90s and led libraries as well as scholars to consider that it would be highly difficult to absorb the ever increasing costs of scientific journals.

The principles of open access have been stated in a wide variety of contexts. The most prominent we can quote is excerpted from the Berlin declaration[2] issued in October 2003 and undersigned by a large number of academic institutions. Open access is presented along two main principles:

- The "free, irrevocable, worldwide, right of access to, and a license to copy, use, distribute, transmit and display open access contributions"

- The fact that "the complete work and all supplemental materials is deposited in at least one online repository using suitable technical standards"

---

1 http://www.pubmedcentral.nih.gov/

2 http://oa.mpg.de/openaccess-berlin/berlindeclaration.html

The first reason why this debate has bearing on our paper here is that the notion of widely accessible information is quite systematically related to that of using open standards and open technologies to represent and disseminate this information.

Secondly, one of the ways people have contemplated the implementation of open access principles has always been to explore and design new publishing models that could somehow be viable alternatives to more traditional commercial publishing. Among such initiatives, pure online journals have been seen as a potentially cheap solution for disseminating scientific information, ranging from pure open access journals like the Living Review series[3], or academic based initiatives (e.g. Revues.org[4]) offering a transitional model for printed journals wanting to move to a digital format.

Finally, one of the main endeavours of the open access supporters, in particular those in favour of the so-called "green" way to open access, is to encourage scientists to deposit their works in publication repositories that freely offer their content (with a possible time embargo) online. Beyond the actual political background, the spread of publication repositories, and most specifically institutional ones, has brought to the fore two important questions that are directly related to the issues addressed in this paper, namely:

- How can the information available in a publication archive, in particular the metadata, may be reused as a reliable source of information for further scientific work?

- How can the model of publication archives be seen as a sustainable one from the point of view of their content, i.e. the capacity to represent full text information in such way that it will still be accessible and legible over a long (digital) period.

As a whole, we claim that some of the technologies and techniques we are reviewing and would like to see take hold will make some types of open-access publishing easier and more effective; but other than that, we will not address the broader debate around open-access any further and in particular aspects related to commercial revenue.

---

3   http://www.livingreviews.org/

4   http://www.revues.org/

## Editing workflows in journal publication

Over the past fifteen years, many thousands of journals have made the transition from print publication to online or hybrid (print and online) publication, without, in most cases, radically changing their authoring and editorial practices. The traditional workflow in the journal publication process involves these stages:

- Submission by author

- Initial decision by editor

- Circulation to peer-reviewers

- Re-editing/rewriting/negotiation between editor and author

- Final editing

- Pre-print / proofing by editor(s)

- Publication

As journal publishing has migrated from print to the Internet, these stages have remained largely intact, and online journal publishing systems have evolved to support them. For instance, the Open Journal Systems documentation[5] describes the OJS editorial process in these five steps:

1. Submissions Queue: Items begin here and are assigned to an editor.

2. Submission Review: Items undergo peer review and editorial decision.

3. Submission Editing: Items undergo copyediting, layout, and proofreading.

4. Scheduling Queue: Items assigned to an issue and/or volume.

5. Table of Contents: Items ordered for publication and issue published.

 Very little has changed here. However, all communications are now mediated through the online journal system rather than through the mail or by telephone; submission is by upload, reviewers access articles through the website, and galleys are proofed through the website. In the OJS system, copyediting and layout are still very traditional; the layout editor creates article files in HTML, PDF or other formats using desktop tools that are not integrated into the online system.

The Public Knowledge Project, parent of OJS, claims that there are "over 2000 titles using OJS (as of January 2009)"[6]. This is a remarkable achievement, and there is no doubt that it has contributed significantly to the large-scale migration of academic journals from print to the Internet. At the same time, many other initiatives have emerged which attempt to take advantage of this transition to re-examine the editorial process. For instance, Blesius et al[7] describe how they created a new electronic publication system for the *Dermatology Online Journal*[8] with a view to allowing users/readers to create and participate in "communities around the content", through online forums, weblogs and other content-sharing tools. Similarly, Copernicus[9], in collaboration with the European Geosciences Union has explored the possibility of introducing community review by means of an open review process, which has proven very efficient in improving the quality of initial drafts and thus augmenting the acceptance rate, with a corresponding reduction in management costs.

---

5  "OJS in an Hour, 2008, p.10,  http://pkp.sfu.ca/files/OJSinanHour.pdf

6  Public Knowledge Project, 2009 http://pkp.sfu.ca/ojs-journals

Unlike OJS (at the time of writing), the *DOJ* publishing system is based on XML, enabling it to "export and share data with external archives using the National Library of Medicine's Journal Archiving and Interchange Document Type Definition." Another journal using an XML-based publication system is the Scandinavian Canadian Studies journal (http://scancan.net/). In this case, the system uses the Text Encoding Intiative (TEI P4 edition). Documents are encoded in XML, and a variety of publication formats are then generated from the base XML automatically, using XSLT transformation; articles are available in XHTML, PDF and plain text format. (They are also available in TEI P5, the successor TEI format, through another XSLT transformation.) The journal still produces a traditional print version, and the PDF document for each full print issue is also automatically generated from the same XML source. One advantage of this is that each article can be proofed and corrected by the editor, the author, and anyone else given access, in the exact form in which it will appear in the final print volume, as soon as it is marked up and injected into the system. In addition, the use of rich markup such as TEI enables automated indexing of any feature that might be included in the markup. For instance, in the case of the IALLT Journal (http://ialltjournal.org/), a system deriving from that used for ScanCan but based on TEI P5 instead of P4, automated indexes are created for all mentions of abbreviations, authors, organizations, people, places, software, and topic keywords. In fact, after several years of publication, the indexes of such a journal will amount to a rich overview of the journal's field, showing who its major and minor figures of significance are, what topics preoccupy it, and what jargon is coming in and out of fashion over time. Another feature of such systems is their elegant handling of corrigenda. An error in an article can be emended as soon as it is discovered, and the change, along with the reasons for it, can be explained in the <revisionDesc> element in the <teiHeader>. The complete set of such errors can be automatically extracted from the database and displayed as a single Corrigenda page.[10] Such features demonstrate clear advantages for a system based on structured markup over one based on print-oriented formats such as PDF or MSWord.

---

7   Blesius et al. "An Open Source Model for Open Access Journal Publication." *AMIA Annual Symposium Proceedings* (2005).

8   http://dermatology.cdlib.org/

9    http://publications.copernicus.org/

10  http://scancan.net/corrigenda.htm

The National Library of Medicine (NLM) XML standards used by DOJ actually constitute a family of standards, with four distinct tagsets, for "Archiving and Interchange", "Journal Publishing", "Article Authoring" and "NCBI Book". In other words, journal articles are intended to be marked-up according to four different DTDs, depending on what is to be done with them. In an extreme case, this might mean:

- The author writes/marks up an article using the Article Authoring DTD.

- Once the article is accepted for publication, the editor or publisher converts it to the Journal Publishing DTD.

- The editor or publisher also creates a version in the Archiving and Interchange format, in order to "supply the content to archives or to interchange it with other organizations".[11]

- The article might also be converted into the NCBI Book format if it is to form part of a textbook.

In reality, the last case — use of a regular journal article directly in a textbook — is unlikely; and the Archiving and Interchange format is intended more for marking up existing print journal content than for use with born-digital articles. All four tag sets are built on the same family of modules, so they do not differ a great deal. Nonetheless, one has to wonder whether the NCBI/NLM goal of "providing a common format in which publishers and archives can exchange journal content"[12] is helped or hindered by the proliferation of variant DTDs.

The online journal Digital Humanities Quarterly[13] also uses a publication system based on its own XML format, "DHQ Markup Language". DHQML also breaks down into DHQauthor (for authoring) and DHQpublish (for publishing). There is a third variant called DHQcrayonbox, which is intended for articles "too 'experimental' for DHQauthor".[14] The authoring variant has the documented goal of being "consonant with tagging constructs familiar from TEI (to the extent possible; processing semantics can take priority but TEI should be used when its semantics fit)," and the DHQpublish schema is intended for "Maximum compatibility with DHQauthor (an easy transform at most)";[15] in other words, these are in some sense variants of TEI.

We can see from this very brief survey that in the field of academic journals, there are now dozens of different formats for online publication; and even in the case of individual journals which might be committed to the use of XML, multiple standards-based or idiosyncratic schemas may be in use.

---

11 http://dtd.nlm.nih.gov/faq.html

12 http://dtd.nlm.nih.gov/

13 http://www.digitalhumanities.org/dhq/

14 http://digitalhumanities.org/view/DHquarterly/SchemaRequirements

15 Op. cit.

## Constituents of journal papers

Before attempting to make any concrete proposal as to the ideal electronic representation of a scientific paper, it is important to have a precise idea about its general organisation, as well as the low-level components such papers may contain. Our aim in this section is thus to identify how much a scientific or scholarly publication departs from any other type of text and, from this, to identify where there is a need for more precise modelling activity for such documents, or at least specific guidelines for applying existing text encoding schemes.

To start with, let us consider the macro-structure of a scholarly article in the generic form it has so far occurred on paper. Independently of any domain-specific restriction or practices, a scholarly paper quite systematically comprises:

- Title of the paper: this comes as the main reference to the scholarly work and usually provides insights on some of the main results, especially in hard sciences:

- Authors, affiliations and addresses: we will come back specifically to this issue later in this section, but here we can point out that author identification information is essential for scholarly work since it provides the basis for the actual attribution of the work to the corresponding researchers. Such factors as ordering or institutional description are here essential in this respect;

- Abstract and keywords: these are intended to provide a means for a quick search in scientific content, in order to select, for instance, those papers which are worth consulting, in the course of a given research project;

- Article body: usually organised in short sections and sub-sections, it typically provides a strong structure that matches closely the main argument of the paper, and may in some scientific domains (e.g. clinical studies) be very standardised in the way certain aspects of the research (methodology, corpus, data gathering, conclusions) are articulated;

- Bibliographical references: another core part of scholarly work since it contains all descriptions of previous scholarly material that were deemed relevant background material for the research presented in the paper;

- Back matter: this comprises a wide variety of small sections such as acknowledgements (to colleagues or research funders), glossaries, appendices (e.g. for data tables, additional graphics, larger quotations) or notes.

At the micro-structure level — that is basically the low-level component of the full-text content — journal papers can be characterized by making systematic use of a few core components that are used in complement to the prose to illustrate, support or formalize the scientific content. Among these, we should pay specific attention to the following ones, which deserve appropriate treatment when represented in a digitized format:

- Bibliographical references: these should be formalized so that, independently of the actual formatting (numbering, author name abbreviation, etc.) we can unequivocally link each citation or reference to an entry in the bibliographical list of the paper;

- Citations: these are structured objects comprising a quotation from a previous work, some possible qualifiers attached to the quotation by the author of the paper (e.g. translation, comment, etc.), and a bibliographical reference to the work. A highly standardized representation of citations would allow many potential overlay applications of bibliographical items across corpora of articles;

- Tables: such components may either be highly structured objects (e.g. numerical data) or purely presentational ones, with possible embeddings. It is necessary to adopt a clear representation policy for tables and assess whether existing schemes (e.g. CALS[16] or XHTML[17]) already match our needs;

- Graphics and images: although they may be considered simple objects, graphics should be treated in a way which is similar to citations, since they may also be associated with comments and bibliographical references about the source. As is the case with tables, existing standards such as SVG[18] provide good options here;

- Mathematical equations, chemical formulae or similar formulaic content: such information may occur either in the course of the plain text or interleaved with paragraphs as block-level items. When not represented as a graphical object, a formula is a highly structured object that requires specific (XML) vocabularies, which should in no case be reinvented by text encoding schemas. For instance, initiatives such as MathML[19] and CML[20] should be used as the basis for the representation of mathematical or chemical content.

At this stage we need to look more deeply at two issues which, from the surface analysis we have just conducted, clearly appear as central in the informational content of a scholarly paper, namely bibliographical references and affiliation information.

First, we would like to make a point of the necessity of having a convergence scenario in mind regarding the representation of bibliographic data, with the objective of ensuring maximal interoperability, but also to anticipate future workflows that will link scientific information across publishers, publication repositories and researchers themselves.

---

16  http://www.oasis-open.org/specs/tr9503.html

17  http://www.w3.org/TR/xhtml2/mod-tables.html

18  http://www.w3.org/TR/SVG/

19  http://www.w3.org/Math/

20  http://cml.sourceforge.net/

As a matter of fact, we should see a continuum in the various bibliographical representations that may occur within or in relation to the paper. The first source of bibliographical data is the paper itself. The digital management of scientific articles indeed requires that precise information related to authors, to the paper itself and to the encompassing journal be recorded in conjunction with the management of the full text. Such information covers aspects, which already existed in the printed world but also information such as author ISSN, DOI or author identification numbers. Secondly, such a metadata description can potentially be seen as the source of future bibliographic information as present in the list of references quoted in the paper. Actually, one or the other level of information should be linked with that available either from publishers themselves (for instance via Crossref[21]) or from publication archives. Finally, such references should not be dissociated from the actual metadata associated with either research data or, in the humanities, with the identification of primary research sources (e.g. corpora), so that, for instance, linking from publications to data and *vice versa* occurs in a homogeneous technical environment.

As a whole, even if some variation may occur from one use case to another (e.g. we may not want systematic affiliation information within a bibliography at the end of a paper), there is a need to design a coherent framework through which all loci of bibliographical data are potentially expressed according to the same principles.

A second important issue, which can be seen as a side aspect of bibliographical representation, has to do with the proper treatment of affiliations. Actually, since the early times of scientific publishing, scholarly papers have always contained information about the authors' organisations and addresses. Initially, such information was intended to provide means for a reader to content an author directly, but this evolved to allow for the precise referencing of the research attribution, when for instance international rankings[22] used this information to assess the research level of academic institutions. Such an evolution created a tension between the necessary conciseness that is required for paper-based affiliation schemes and the precision that is expected to provide a sound basis for research attribution activities.

The transition to digital publishing somehow resolves the dilemma by offering a different perspective on both the management and representation of such author-related information. As a matter of fact, one of the underlying difficulties is that, so far, most bibliographic or bibliometric databases have used the printed version of a paper to extract affiliation information. Providing a born digital version of a paper with precise author-related information permits publishers to provide a reliable source, which can then be further consumed by information integrators. It should be noted here that publication archives can also, when managed by academic institutions, be a reliable source for such affiliation data.

---

21  http://www.crossref.org/

22  e.g. Shanghai ranking of Universities, see http://www.arwu.org/

A direct consequence of this is that digital formats for journal article archiving, as well as for all steps in the editorial workflow, should be designed in such a way that they can express a fine-grained representation of authors' affiliations and addresses. In this respect, it is probably a mistake to design such a format by mimicking the paper representation of authors' addresses[23] as coindexed to us concerning author reference actual rather than providing an integrated representation. This is again  an opportunity for convergence, where a systematic approach to the digital representation of affiliation is aimed at in the context of a digital journal scenario.

## XML formats: what are the options?

Now that we have identified the main components of a journal paper, we can have a closer look at the options opened to us concerning their actual digital representation. Still, it is hardly possible to make an actual choice or even to have a global vision unless we situate the perspective of the representation of journal content within some basic use cases pertaining to the journal workflow, namely editing, publishing and archiving.

At the editing stage, the emphasis is basically to offer the best compromise between the flexibility required by author in providing their manuscripts and the editorial coherence that the journal may want to impose across all its published content. Since the corresponding draft may not necessarily have a long lifetime, standardisation constraints are rather low, even if great attention should be paid to processes allowing content validation and checking (affiliation, bibliography, coherence of internal references to figures, tables and graphics). The actual format to be used internally for this editorial stage may also depend on the capacity to be interoperable with the various platforms and software potentially used by authors and journal editors.

The publishing stage introduces a set of somehow reverse constraints from the editing stage. The emphasis is indeed here to move from one reference version of the journal paper to a multiplicity of potential presentational formats, such as the creation of a printed version (if applicable), the production of an online distribution version (e.g. in pdf), the setting of a (possibly reduced) consultation format in html, as well as the generation of various output versions to feed the journal's webpages (title, author and summary for instance), or various databases such as Crossref. This requires that the underlying format be structured in such a way that filtering out and reorganising its content can be fully automated and combined with a variety of layout structures.

Finally the archival stage is intended to ensure long-term reusability of the journal content both by humans (legibility) and/or machines (processability). We should also distinguish here between the aspects of preservation, and availability for re-use. For instance, a PDF document is well suited to preservation, since it is likely that PDF display and printing software will be widely available for a long time in the future, and the original print form of the document will be accurately represented through such means. However, it is not easy to take a PDF document and re-purpose it. Text, when extracted from a PDF, is in block-fragments (usually lines), and is organized by physical position on the "page"; it has no conceptual or hierarchical structure, and cannot easily be transformed into another kind of document. When considering what might constitute an appropriate format for archiving, it is well to consider whether we are attempting to archive its physical representation (in which case a series of TIFF images of the printed pages, or a standard PDF document would presumably suffice), or its conceptual structure and content (in which case we should be looking for a format which encodes the hierarchical/structural organization of the document, and identifies its constituents according to what they are rather than what they look like (e.g. a book title, rather than a span of italicized text).

---

23   See the NLM proposal as a good example of this strategy.

At this stage, we want to support and explore further the hypothesis that it is necessary to work towards a back-office representation of journal papers that can seamlessly take into consideration the constraints of the editing, publishing and archival stages. In addition, we do think that such a format, or family of formats, should also be integrated within a wider perspective of interoperability (whether partial or total) with, on the one hand, other textual documents (reports, research notes, primary sources, glossaries) and, on the other hand, with other forms of scientific outputs. The perspective adopted here is indeed not far from the notion of *datument* advocated by P. Murray-Rust and H. S. Rzepa[24].

The next stage for us is to look at the various existing formats and see how they match the constraints identified so far. As to the current practices, textual documents are mostly deposited in the formats that have been used for their editing or human oriented dissemination. These fall into three main categories:

- Tex/Latex-based source documents, which are used in specific scientific communities (e.g. Mathematics, physics, computer science) and are compiled to produce a legible Postscript or PDF output. The possibility to define specific mechanisms through macros results in a high variation in the actual expression of document structure and content;

- Word processing proprietary files, which are dependant on the actual piece of software and version thereof. This dependency creates an important problem as to the long-term sustainability of the corresponding documents;

- Presentational formats such as Postscript and (now mainly) PDF, which have been designed by private companies. A specific version of PDF (PDF/A) has been stabilized as an ISO standard[25] dedicated to the provision of a long-term archiving format for electronic document at large. As we mentioned above, while it is likely that software for reading, displaying and printing PDF documents will be available for the foreseeable future, it is quite difficult to edit PDF documents, and even more difficult to transform them into more conceptually-structured formats.

This situation has developed in parallel to the wide spread of the XML recommendation, which provides a generic framework for the representation of digital objects, and which has soon been considered (or even strongly advocated, see Murray-Rust and Rzepa, 2003) as the unavoidable basis for a long-term archival strategy of publication documents. Arguments in favour of adopting XML can be easily summarised as follows: it is based on a simple formalism yet offering a good expressive power (tree structures), is straightforwardly legible, which is essential in a long term archiving perspective, and its wide dissemination has not only yielded a wide range of generic tools, but also specific reusable components (XLink, CALS, MathML and the like) that provide local interoperability across applications.

As a matter of fact, once the reference to XML is made, we should immediately point out that the stable syntactic framework it provides is not enough to guarantee full interoperability. Beyond the syntax, it is essential to consider that one also has to share dedicated vocabularies and the corresponding semantics. In the perspective of journal papers, this relates to the issue of identifying how much coverage we have of the various components that we identified earlier in this paper.

Indeed, the situation in this respect is still rather fragmented and has not led to a clear strategy to crystallize an XML-based format for scientific publications which would be minimally suited for long-term archival. In fact, there are currently several potential candidate endeavours:

- XML formats related to word processing platforms, mainly the OpenDocument format (ODF; developed in the context of Open Office) and Office Open XML (OOXML; by Microsoft), both of which have gone through an ISO standardisation process. Their relation to editing processes and thus to the presentation of content prevents them from being used as archival formats. In particular they both bear a high complexity specifically linked to the nature of word processing.

- Highly specialised XML formats dedicated to scientific publishing activities, either within specific publishing or archival initiatives (Erudit[26]) or created in relation to archival initiatives (DiVA). The NLM family of formats, which we addressed previously, also falls into this category;

- Generic XML formats targeted at the representation of the logical content of textual documents. The two main relevant initiatives in this respect are DocBook and the TEI, which both provide a rather large spectrum of encoding possibilities while preserving a generic document structure applicable beyond the sole case of scientific publications.

---

26  http://www.erudit.org/

Furthermore, the TEI is organized as an international consortium, which provides a wide base of expertise for the maintenance and improvement of the guidelines. From a technical point of view, and beyond the more than 500 elements it already contains, the TEI offers a framework where it is possible to design specific customisations while remaining compliant with the guidelines as a whole. This is particular important in a context where specific editorial projects related to certain scientific fields may need to express their own constraints. This is also a way to avoid the necessity to design, right from the outset, a specific format for authoring, archiving or publishing purposes. In this context, whereas DocBook or NLM could be seen as good candidates for representing journal content, we think the TEI offers potentially a larger, more broad-based and generic standard than any of them. Beyond the possibility to actually share more tools and technical settings, the TEI brings in a conceptual framework, which can be shared with a wider community than those strictly interested in the representation of scholarly papers.

## Creating a new standard

In our discussion above, we have argued for the desirability of a single unifying journal mark-up schema, which could be used by a majority of electronic journals, at least within the Humanities; and we have suggested the TEI as a good candidate to form the basis of such a schema.

The Text-Encoding Initiative[27] has been developing and documenting schemas for the digital humanities community for more than 15 years. The current version of the TEI schema, P5, is a complex and very sophisticated set of modules comprising many hundreds of elements and attributes. Historically, TEI has been used primarily to create digital encodings of existing historical texts. In recent years, however, it has increasingly been used to create born-digital content.[28] For instance, the DHQ schemas discussed above are actually based on TEI. As an encoding format for scholarly publications, TEI has many advantages:

- As mentioned above, it is already well-tuned for the markup of existing physical documents, so older print articles can easily be migrated into TEI.

- It has a range of modules specifically designed for addressing the needs of humanist scholars (specialized tags for use with manuscripts, for handling obscure languages and linguistic features, etc.).

- It already integrates well with many existing standards and schemas such as SVG (for vector graphics), MathML (for formulae etc.), W3C and ISO date formats, XHTML (for tables), and so on.

---

- It is designed from the ground up to be customized for specific purposes, and comes with tools for creating, documenting, publishing and using customizations.

- There is a large community of existing TEI users, as well as a large base of existing texts and projects.

We believe that creating a journal article schema framed as a TEI customization would enable us to strike a balance between these three components:

- Prescription: encouraging encoders to adopt specific practices which the community feels are effective and appropriate.

- Arbitration: selecting and endorsing one approach (or a small number of approaches) to a specific encoding requirement, in the interests of formal simplicity, interoperability and uniformity.

- Codification: formal schematization of what encoders already actually do.

## Outline of a TEI-based schema for representing journal papers

It would obviously be beyond the scope of this paper to provide a fully-fledged description of what a TEI customization for scholarly papers could be. Still, we would like to point to a few aspects where clear recommendations could be made, and, doing so, demonstrate the capacity of the TEI guidelines to cover some of the core features that we deemed essential for this textual genre. Starting with an overview of an article macro-structure we will point out specific mechanisms, in particular in the domain of bibliographical representation that are particularly relevant for journal paper encoding.

### General structure of a TEI document

The TEI information model is intended to represent both the textual content of a document and the metadata attached to it. This is reflected in the two main parts of a <TEI> root element, namely <teiHeader> and <text>.

The TEI header is in turn organised in a series of sub-components:

- <fileDesc> gathering the main characteristics of the document (title, author, bibliographic description of the source). This is the main place where metadata information will be expressed (see below);

- <profileDesc> providing some information about the content. This is the place where such information as the languages used in the text or the provision of keywords (see example) should be situated;

- <revisionDesc> providing the history of the document. In the context of an editorial workflow, this should be used to trace the history of the paper (submission, review, revision, publication).

The <text> element is further decomposed into <front>, <body> and <back>. When available, abstracts are represented in <front> and full-text content in subsequent elements.

### Skeleton of a full TEI document

We present below a model structure of a TEI document as we would see it relevant for the representation of a journal paper. Such a skeleton already reflects a few issues where specific implementation choices have been made, namely:

- The use of <biblStruct> in <sourceDesc> as the sole structure to represent bibliographic data attached to the paper;

- The duplication of the article title in <titleStmt> to facilitate interoperability with other types of TEI documents when put together, for instance, within a digital object management system;

- The insertion of copyright information in <publicationStmt>;

- The representation formats for keywords attached to the paper;

- The use of <revisionDesc> for tracing the editorial stages of the paper.

```xml
<TEI xmlns="http://www.tei-c.org/ns/1.0">
    <teiHeader>
        <fileDesc>
            <titleStmt>
                <title level="a" type="main">...</title>
            </titleStmt>
            <publicationStmt>
                <availability>
                    <p>Copyright © The Animal Consortium 2009</p>
                </availability>
                <date>2009</date>
                <authority>The Animal Consortium</authority>
            </publicationStmt>
            <sourceDesc>
                <biblStruct>...</biblStruct>
            </sourceDesc>
        </fileDesc>
        <profileDesc>
            <textClass>
                <keywords>
                    <list>
                        <head>Keywords</head>
                        <item>
                            <term>foetal development</term>
                        </item>
                        <item>
                            ...
                        </item>
                    </list>
                </keywords>
            </textClass>
        </profileDesc>
        <revisionDesc>
            <change when="2008-08-27">Received</change>
            <change when="2008-12-01">Accepted</change>
        </revisionDesc>
```

```xml
    </teiHeader>
    <text>
        <front>
            
                <head>Abstract</head>
                <p>...</p>
            
        </front>
        <body/>
        <back/>
    </text>
</TEI>
```

## Representation of bibliographical information

As stated earlier, the representation is based on the TEI <biblStruct> element, which is organised as follows:

```xml
<biblStruct type="article">
    <analytic>
        …
    </analytic>
    <monogr>
        …
        <imprint>
            …
        </imprint>
    </monogr>
    …
</biblStruct>
```

A <biblStruct> is mainly divided into two sub-structures:

- <analytic> indicates the bibliographical characteristics of an article (title and authors);

- <monogr> accounts for the publication details of the journal (journal name, publisher information, issn, etc.), and contains in turn a <imprint> element which gathers publication and/or distribution aspects of the article in the corresponding journal (pagination, volume, issue, etc.);

- When applicable, additional notes or identifiers can follow, for instance, the DOI, PubMed Central id or repository-specific id will appear here:

```xml
<biblStruct type="article">
    <analytic>…</analytic>
    <monogr>…</monogr>
    <idno type="pmid">12345678</idno>
</biblStruct>
```

## The <analytic> element

The title of a journal article is represented by means of the <title> element (with appropriate @level attribute) as follows:

```xml
<title level="a">Multilocus Analysis of Age Related Macular Degeneration</title>
```

When necessary a further @type attribute may be used to differentiate between main and subtitles (@type="main" vs. @type="subordinate"), as well as specific titles such as recto and verso running titles (at publication stage).

Each author in the <analytic> element is independently described by means of an <author> element. This element contains the author's name, affiliation and addresses − when available − together with some possible generic author identifiers[29] as presented in the outline below:

```
<author>
    <idno type="...">...</idno>
    <persName>
        <forename>Michael</forename>
        <surname>Dean</surname>
    </persName>
    <affiliation…</affiliation>
    <email>dean@ncifcrf.gov</email>
</author>
```

The <affiliation> component of <author> is intended to contain any potentially relevant information with regard to the author's academic situation: research group, laboratory, institution.

```
<affiliation>
    <orgName type="laboratory">CSA Department</orgName>
    <orgName type="institution">Indian Institute of Science</orgName>
    <address>
        <settlement>Bangalore</settlement>
        <postCode>560012</postCode>
        <country>India</country>
        <addrLine type="phone">+91-80-22932386</addrLine>
        <addrLine type="fax">+91-80-23602911</addrLine>
    </address>
</affiliation>
```

Such a representation provides a clear way of identifying, in a standardized manner the various organisational levels to which a research may be affiliated. Further standardisation would typically include defining precisely the permitted values of the @type attribute on <orgName>, at least in the context of contextual (regional) research organistion schemes, or in relation to classification scheme adopted by major vendors such as Thomson scientific with the Web of Science.

### The <monogr> element

The <monogr> element gathers journal identification information (journal title and ISSN together with the publishing information contained in its <imprint> sub-element). For instance:

```
<monogr>
    <title level="j" type="main">European Journal of Human
Genetics</title>
    <title level="j" type="nlm-ta">Eur J Hum Genet</title>
    <idno type="ISSN">1018-4813</idno>
    <imprint>…</imprint>
```

```
</monogr>
```

**The <imprint> element**

"By imprint is meant all the information relating to the publication of a work: the person or organization by whose authority and in whose name a bibliographic entity such as a book is made public or distributed (whether a commercial publisher or some other organization), the place of publication, and a date. It may also include a full address for the publisher or organization. Full bibliographic references usually specify either the number of pages in a print publication (or equivalent information for non-print materials), or the specific location of the material being cited within its containing publication."[30]

The <imprint> element is organised as follows:

```
<imprint>
    <pubPlace>Oxford</pubPlace>
    <publisher>Clarendon Press</publisher>
    <date typ="published" when="1969-02-07"/>
    <biblScope type="vol">3</biblScope>
    <biblScope type="issue">2</biblScope>
</imprint>
```

The possible values for the attribute @type on <biblScope> are the following:

- vol: volume

- issue: issue

- fpage: first page

- lpage: last page

- pp: number of pages when the information about full pagination is not available[31]

**<biblStruct> skeleton**

The following example provides an overview of the full internal structure of the <biblStruct> element as suggested for the standard representation of bibliographical information attached to a journal paper:

```
<biblStruct type="article">
    <analytic>
        <title level="a" type="main">…</title>
        <author type="corresp">
            <persName>
                <forename>…</forename>
                <surname>…</surname>
            </persName>
            <affiliation>
                <orgName type="">…</orgName>
                <address>…<country>FR</country></address>
            </affiliation>
```

---

30  http://www.tei-c.org/release/doc/tei-p5-doc/en/html/CO.html#COBICOI

31  We restrict here the semantics of the recommended value (cf. http://www.tei-c.org/release/doc/tei-p5-doc/html/ref-biblScope.html)

```
            <email>…</email>
        </author>
    </analytic>
    <monogr>
        <title level="j" type="main">…</title>
        <idno type="ISSN">…</idno>
        <imprint>
            <publisher>…</publisher>
            <pubPlace>…</pubPlace>
            <date when="2009-02-03"/>
            <biblScope type="fpage">…</biblScope>
        </imprint>
    </monogr>
    <idno type="DOI">…</idno>
</biblStruct>
```

### Consequences for article micro-structure

As can easily be seen, the bibliographical format presented above is generic enough to cover all needs for structuring inline bibliographical references. Basically, this would correspond to exactly the same structure with possible simplifications regarding author affiliation. As elucidated in the TEI guidelines, the <biblStruct> element actually covers a wide range of bibliographical types ranging from conference papers to books and can impact at two major places within a journal paper:

a. In the list of bibliographical references of a paper, which can be very uniformly represented as a <listBibl> of <biblStruct>s;

b. In inline citation, for which the TEI typically offer a generic construct outlined in the following example where one can see how precise bibliographic reference can be association with the quoted text:

```
<cit>
    <quote>Wer A sagt, der muß nicht B sagen. Er kann auch erkennen,
    daß A falsch war</quote>
    <biblStruct>
        <monogr>
            <author>
            <persName>
                <forename>Bertolt</forename>
                <surname>Brecht</surname>
            </persName>
            </author>
            <title>Der Jasager und der Neinsager - Vorlagen, Fassungen und
            Materialien</title>
            <imprint>
                <publisher>Edition Suhrkamp</publisher>
                <date type="Published" when="1981"/>
            </imprint>
        </monogr>
        <idno type="ISBN">9783518101711</idno>
    </biblStruct>
</cit>
```

Without going any further here in the precise description of TEI mechanisms, we hope we have made it clear how the TEI guidelines could match the needs of scholarly publishing by providing generic mechanisms which can in turn be tuned (probably with additional recommendations) for journal papers. The next step for us is to identify how to articulate these facilities with the actual design of a journal publishing workflow.

## Approaches to creating a TEI-based schema

In the sections above, we have given some suggestions as to how an ideal schema for journal markup could be based on TEI. We might call such a schema "teiJournal". Most likely, this would be a stripped-down form of TEI, meaning that customization would consist only of the application of constraints: in other words, a teiJournal document would be fully TEI-compliant (meaning that it would validate under the "full" tei_all schema which incorporates all the available modules in TEI). There is considerable value in this; a fully TEI-compliant schema provides instant interoperability with any system that understands TEI. At the same time, the requirements of a journal schema are considerably restricted compared with the huge range of needs that the TEI itself attempts to answer. For instance, since a journal schema would be used primarily to encode born-digital documents, it might not require many of the TEI elements and attributes related to (for instance) manuscript description, or "certainty and responsibility". At the same time, we would expect that a teiJournal schema would be more prescriptive than the general TEI Guidelines with regard to certain specific encoding problems. For instance, most journal articles include some kind of sources list or bibliography, and it would be a primary requirement of any processing engine that such a list be rendered into a highly formalized output format, conforming to the prescriptions of a style guide such as MLA, APA or Chicago. In order to do this, a highly-structured markup format would be required, and we have argued that the TEI <biblStruct> element would be most appropriate for this task, so the looser <bibl> and <biblFull> elements which TEI also provides for different usage scenarios could be discarded from the schema in the interests of simplicity.

At this point, we will look at a primary requirement of any journal publishing engine: to render different types of document in different ways, as prescribed by the various style guides in use in the academic publishing realm. For instance, when rendering the content of an article's bibliography in XHTML or PDF for the end user, journal titles may have to be italicized, while article titles should appear in quotation marks. From our previous work designing applications to render bibliographical lists like this[32], we have identified at least 60 different types of document[33] which may need to be handled in different or idiosyncratic ways by a rendering system in order to comply with the differing requirements of the various style guides. A natural way to distinguish different types of document would be to use the @type attribute on the <biblStruct> tag:

```
<biblStruct type="book">...</biblStruct>
<biblStruct type="journalArticle">...</biblStruct>
```

The TEI Guidelines say that @type "characterizes the element in some sense, using any convenient classification scheme or typology," and its type is data.enumerated; "Typically, the list of documented possibilities will be provided (or exemplified) by a

---

value list in the associated attribute specification, expressed with a valList element." The problem then is generating this value list, and typically this would involve a process of trying to predict every possible required value, and negotiate an agreement on the exact form of each. Any attempt to create a standard will inevitably expend a great deal of time and effort on devising and refining feature lists such as this, and the results are rarely completely satisfactory; no sooner is a standard released than real-world users discover needs that the standard cannot yet accommodate.

However, a recent contribution to the TEI toolset by Sebastian Rahtz has opened the way to a new approach we might take to solving problems like this. Rahtz has released an XSLT transformation called "oddbyexample.xsl" which is designed to "read a corpus of TEI P5 documents and construct a ODD customization file which expresses the subset of the TEI you need to validate that corpus."[34] (An ODD file is an XML file which expresses the details of a TEI customization: which elements and attributes from the overall TEI system are included and excluded from the schema, and what their values and behaviour might be.) This tool has the potential to allow rapid generation of restricted TEI schemas based on a corpus of documents — essentially an approach based entirely on "codification" as defined above. Using oddbyexample.xsl, we can generate a "tight" schema from a collection of documents, and then validate new documents against that schema. We can now consider a much more bottom-up, community-based approach to the generation of a teiJournal schema, which might work like this:

1. A group of users concerned with using TEI to encode journal articles agree to work initially with a large TEI schema -- perhaps even tei_all, but most likely a version with some irrelevant modules removed.

2. They agree on some basic rules (overall document structure, use of <biblStruct>, etc.).

3. They begin encoding. Each completed document is submitted to a central corpus.

4. At a certain point, oddbyexample.xsl (or something similar) is run against the corpus, generating a very stripped-down schema. At this point, all completed documents will validate against this schema; it represents the range of what encoders are actually doing.

5. The community can examine this schema, and look specifically for places where more than one competing approach is being taken to the same encoding issue. To take a trivial instance, perhaps some people are using <hi rend="italic"> and others are using <hi rend="italics">, because the content of the @rend attribute is not restricted in the standard TEI schema. The new, generated schema will provide an enumeration as the content of @rend, but that enumeration will be based on what has been used, so both "italic" and "italics" will be permitted; this is clearly not a desirable situation. A decision can be made to standardize on one of these, or on something else (perhaps <hi rend="font-style: italic;">). Then all existing documents are converted to use the standard format, and a new schema is generated using oddbyexample.

6. All future encoding proceeds based on the new, restricted schema. Then, when a novel need arises — someone needs to encode something which is not handled by the schema — they can simply switch back to the original TEI schema, and use elements and attributes from the larger set.

7. Periodically, oddbyexample is run on the corpus again. Any elements and attributes from the larger set which have been incorporated in new documents will now find their way into the restricted schema, which grows a little based on need.

Over two or three years, assuming enough encoders and projects are involved with this project, a tight but powerful schema should emerge from a process like this. In addition, the work itself is less time-consuming and stressful than a traditional working-group approach, since the schema emerges naturally over time, and encoders are able to proceed with their projects throughout. The only minimal disruption would be the occasional necessity to transform existing documents whenever "arbitration" takes place to select one approach out of several that are in use. XSLT should be able to handle most such cases.

Once again, we can see how, with its built-in support for schema customization, TEI is particularly suited to schema-development that proceeds in such an "evolutionary" manner, because TEI has such a wide range of existing elements, attributes and encoding strategies from which the process can draw whenever there is a need to handle a new feature.

## Pros and cons of using distinct flavours of the schema for authoring and publication.

One question that should be addressed is the issue of distinct schema variants for different purposes. It is notable that both DHQ and NLM have one schema for authoring, and one for publishing. It is worth quoting at length from the explanation on the NLM website explaining how the authoring schema differs from the publishing schema:

> The Article Authoring Tag Set creates a standardized format for new journal articles that can be used by authors to submit publications to journals and to archives such as PubMed Central. While in theory the document scope is the same as for the Publishing Tag Set, in practice Authoring defines elements and attributes that describe the content of typical research-style journal articles.

> This is a Tag Set optimized for authorship of new journal articles, where regularization and control of content is important, and where it is useful rather than harmful to have only one way to tag a structure. Therefore Authoring is more prescriptive than descriptive and includes many elements whose content must occur in a specified order.

> Since an author is assumed to be creating and submitting an article for submission to a journal or journals, no publishing history or journal-specific information has been included in this Authoring Tag Set.

> Since no assumptions can be made concerning the processing software or editorial situation that will receive an article authored in this Tag Set, tagging that forces specific formatting has also been avoided. There is no way for an author to number his/her lists explicitly, for example, or to manually number the cited references, since many journals have their own citation policies and publication styles. Numbers for the cited references must be generated by the publisher's software to match editorial policy and established practice.[35]

In fact, in practical terms, the differences are of minor importance; in the case of the three example marked-up documents provided on the NLM website as part of the tagset documentation (two "publishing" and one "authoring"), all three validate under both schemas. In an additional test, we took nine sample documents converted from a TEI schema to the NLM publishing tagset as part of another project, and successfully validated all nine under both the publishing and authoring schemas. The case of DHQ seems to be very similar; the documentation for the DHQpublish schema suggests that the only difference is that "the DHQheader element is required, and contains a superset of the elements allowed in the DHQauthor header."[36] Also, the sample DHQ-MonkeyHouse.xml document provided for users of the DHQ schemas also validates under both schemas, with the sole exception of a missing <publicationStmt> element in the header, and this turns out to be in the document, but commented out; when included, the document validates under the publication schema but not the authoring schema, and when excluded, vice versa. DHQ does complicate the process a little more, actually, by the provision of two root tags for authoring:

> The document element or "root element" of a DHQauthor document will be either DHQdraft or DHQarticle. The only difference between them is that in DHQdraft, the DHQheader element is optional. You can encode your article using the DHQdraft element to begin with, but all articles submitted to DHQ must use the DHQarticle structure and must include a DHQheader.[37]

Frankly, this seems like unnecessary complexity, since even if the author starts off using DHQdraft, the document will have to be converted to DHQarticle before submission anyway.

So the distinction between authoring and publishing schemas is apparently trivial, and appears to be an attempt to be kind to authors, avoiding distracting them from their work by intruding aspects of publication formatting and metadata into their authoring process. However, we have already noted the tendency for authors writing for modern online journals to be more involved in the markup and layout process;[38] in a sense, many authors are now full participants in the construction of the published artefact. They will imagine their contributions in publication form, and proof them in something approaching it. So why remove publication-related markup features from their schema? Editors may surely edit markup just as easily as text, and the final decision on all aspects of an article lies with the editor, but there is no reason to prevent an author from contributing to the creation of publication metadata, layout decisions, and other aspects of markup currently reserved for the publication schema.

---

35  http://dtd.nlm.nih.gov/articleauthoring/

36  (DHQpublish RelaxNG schema, version "beta", October 2007)

37  http://digitalhumanities.org/view/DHquarterly/TagLibrary

38  See Blesius et al, 2005, and also the editorial process of the *Scandinavian Canadian Studies Journal*, discussed above, in which authors proof their documents through the publication engine, seeing them in the exact form they will appear when published.

Another distinction maintained by NLM is that between new, straight-to-NLM content, and documents intended for archiving and interchange. The Archiving and Interchange tagset "enables an archive to capture structural and semantic components of existing material without modeling any particular sequence or textual format" (http://dtd.nlm.nih.gov/archiving/). This aim is, on the face of it, similar to some aspects of the TEI's purpose: to preserve in digital form material which was originally created in print or some other analogue format. (However, the TEI of course goes further, allowing for as much descriptive information as possible about the original document to be captured along with its structure and semantics.). For such a markup schema, there is no particular inherent output target or intended processing engine. This aim is largely irrelevant to the current discussion, because it is essentially preservative, while markup for born-digital publication is essentially original and creative. Also, in the case of digitizing old content or converting other formats for archive, we are no longer interacting with the content and changing it. Modern online journals, by contrast, appear to be evolving in the direction of greater involvement on the part of a larger number of interested parties — authors, editors, readers, reviewers, collaborators, commenters — all of whom potentially affect the evolution of a published piece. However, it is worth noting that the TEI's origins and primary function make it peculiarly suited to the digitization of existing print content, and it would be perfectly practical to mark up a historical article using TEI such that it would conform to a teiJournal schema (and thus be manageable by a publication content engine), while at the same time including all the descriptive information that a traditional digitization project would wish to record about a historical document. The TEI can perform both functions simultaneously.

It seems, then, that we should be able to settle on one schema for born-digital content, and stick to it, rather than elaborating the system with variants for authoring, editing, archiving and so on. If we have the desire to avoid distracting authors by the inclusion of editorial publication features in a schema they will use, then we can certainly create a more stripped-down variant, supply default placeholder values in a skeleton document, or use some similar mechanism to achieve the same aim. After all, an authoring schema that produces only documents which validate under the publication schema — which is simply a subset of the publishing schema — is arguably not really a different schema at all. Whether it makes sense to do this at all is another question. In the case of the DHQ schema, for instance, is it really that distracting for the author to encounter the need for a publicationStmt tag:

```
<publicationStmt>
  <idno type="DHQarticle-id">001</idno>
  <idno type="volume">1</idno>
  <idno type="issue">1</idno>
  <issueTitle>Summer 2008</issueTitle>
  <articleType>article</articleType>
  <date when="2008-07">July 2008</date>
  <availability>
```

```
    <cc:License rdf:about="http://creativecommons.org/licenses/by-nc-
nd/2.5/"/>
    </availability>
</publicationStmt>³⁹
```

and have to supply some default values? It's certainly no bad thing to be reminded that your article will be released under such-and-such a Creative Commons licence, and it's hardly confusing to know that it will eventually have a volume and issue number, and a publication date.

## Implementing a publication engine

Given the choice of XML as an encoding format, a wide range of tools for storage, retrieval and delivery of content are available. For a back-end storage engine, it could be said that almost anything will do, since XML is "text" and can be queried through any traditional text query engine. However, as we have seen, among the great strengths of XML are its hierarchical structure and conceptual tagging, and a good publication engine should be able to take advantage of these features, both for querying and searching, and for delivery of the content in a variety of different forms. XQuery (XML Query Language) is the natural way to do this; it was designed specifically for precise searching, extraction and restructuring of XML data. Increasing numbers of conventional relational database engines, including Oracle and Microsoft SQL Server, and are now adding support for XML through implementation of interfaces based on XQuery. Another class of database includes "pure" XML databases such as the open-source eXist,⁴⁰ in which data is stored not in a set of two-dimensional tables with rows and columns, but in a "collection", which consists of nested subcollections of documents, together constituting a single XML hierarchy, with an index of every single tag and attribute in the hierarchy.

Once data is extracted through XQuery – whether a complete document, a small fragment of a document, or a collection of related fragments from across the collection – it must be formatted for delivery to the end-user. XML itself is not really an end-user format; although it can be quite attractively styled with the direct application of CSS, such a simple delivery mechanism is unlikely to be full-featured enough, since it will lack features such as hyperlinking and interactivity. More commonly, the content will be transformed, through the use of XSLT, another XML standard language, whose purpose is to convert XML structures into other types of output. Typical output targets will be XHTML (for display in a browser), PDF (for printing, or display in an eBook reader), and perhaps also plain text (for input into text analysis engines, or for a Project Gutenberg-style electronic text). Production of a PDF is typically a two-stage process, in which the initial XSLT transformation creates an XSL:FO document, which is then transformed by a PDF generator engine such as XEP⁴¹ or FOP⁴² into a PDF or PostScript document.

---

39 DHQ-MonkeyHouse.xml sample document,
   http://digitalhumanities.org/twiki/pub/DHquarterly/DownloadCentral/DHQ-MonkeyHouse.xml

40 eXist Open Source Native XML Database, http://exist-db.org/

41 http://www.renderx.com/tools/xep.html

The diagram in Illustration 1 demonstrates the process described above as it is implemented in the case of the *Scandinavian Canadian Studies* journal[43]. On the journal web site, XHTML, plain text, PDF and XML versions of the journal articles are available, all generated on-the-fly from the XML database; the list of contributors with their biographies is also generated from the XML collection, and the search system queries the same system to retrieve document fragments as "hits". The print version of each issue of the journal is generated from the same XML source documents via a more complex XSLT-to-XSL:FO-to-PDF transformation which automatically generates the Table of Contents, indexes, page numbering and so on, with the final stage being accomplished by the commercial XEP engine (although open-source PDF generators such as FOP are available too).

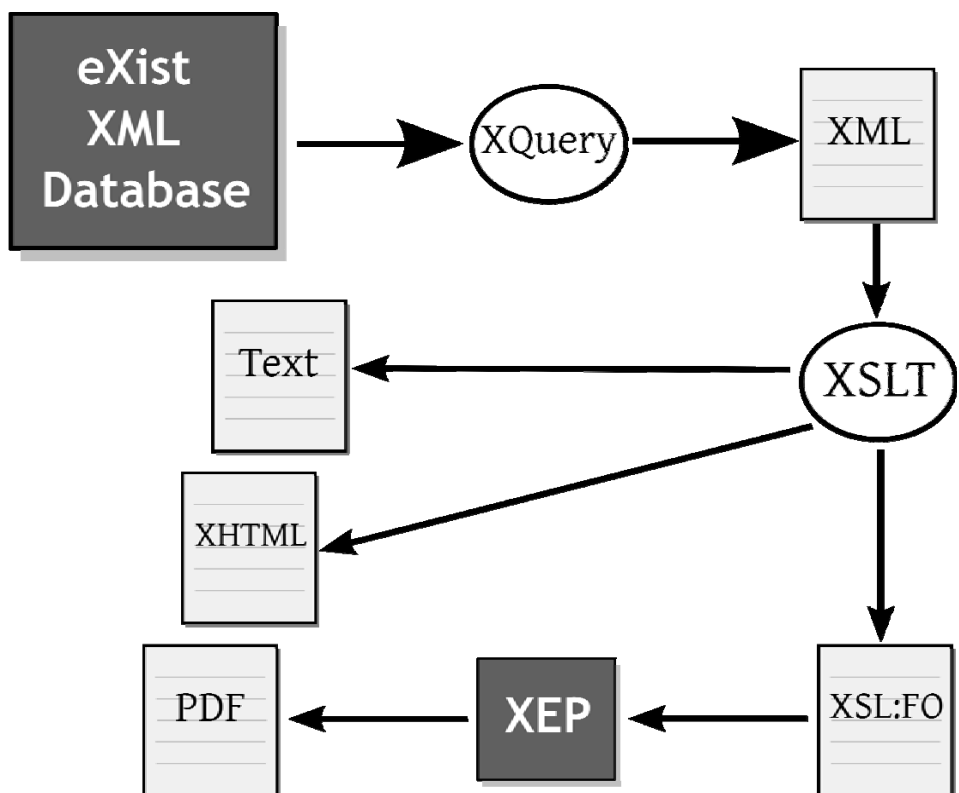

*Illustration 1: The publication engine of the Scandinavian Canadian Studies journal*

This is the solution to the dilemma posed by Thom Lieb in his 1999 article "Q. A.: HTML, PDF and TXT: The Format Wars".[44] Lieb's brief article concludes thus: "The ideal for many online publications would be a combination of all three: a plain-text e-mail alert, an HTML version for fast loading and online reading, and a downloadable PDF version for offline reading." We can now provide all three from the same source. Among the many advantages of a system like this are these:

- As new export formats come along, new output paths can easily be added to the system, generating new document types from the same source. Transient Web format-fashions can easily be added to the system – creating an RSS feed of titles and abstracts of articles as they are published, or Twitter "tweets" announcing new articles would be simple tasks, and such features can be turned off when their moment has passed.

- If it becomes necessary, the whole document collection can be migrated (via XSLT) to another format/schema, and inserted into a different publication engine.

- If it is desirable to serve this content through a system such as an older OJS install, which requires a static file in (for instance) PDF format for each article, the PDF output from the XML-based system can simply be injected into the other engine.

- Authorial and editorial practices, as well as compliance with a styleguide, will be built into the system at the level of the XQuery and XSLT operations, so changes to these systems can be made in a single centralized location, and immediately apply to all the articles in the system. For instance, if the editorial board decides to change the journal styleguide from APA to Chicago, the documents themselves will not need to be changed; only the output transformations will need to be revised. In a system in which documents are stored in a static format such as MS Word, such a change would require re-editing of all the existing journal articles.

- The collection can be treated as a single composite source document, so for example a unified bibliography can be compiled automatically from all the references in all the documents. This has obvious scholarly value.

- The nature of XML tagging in a schema such as TEI allows for highly sophisticated search systems which target specific tags at particular locations in the hierarchy. For example, you could limit a query so that it searches only inside the names of individuals, or the names of organizations; or you could search for all the documents published within a particular date range whose bibliographies list works by one specific author.

## Conclusion

In this article, we have argued that there is a strong need for a single standard format for scholarly and scientific articles, and that current "archive" formats such as PDF and DOC are unsuitable for this purpose; XML is a better option. We have further proposed that, despite the fact that at least two existing XML standards (NLM and DocBook) are already in use for this purpose, a format based on the Text-Encoding Initiative schema would be a better alternative for a variety of reasons. We have given some details of what a TEI-based document structure for journal articles might look like, and examined some of the specific encoding issues that are particularly relevant to sphere of scholarly journals, and we have outlined a bottom-up, rather than top-down, procedure in which the TEI community might be able to evolve a new standard, rather than striking a committee to sit down and devise one. Finally, we have looked at the kind of publication engine that can be built around an XML document collection, and outlined some of its advantages.